\documentclass[conference]{IEEEtran}
\IEEEoverridecommandlockouts
\usepackage{cite}
\usepackage{amsmath,amssymb,amsfonts}
\usepackage{algorithmic}
\usepackage{graphicx}
\usepackage{textcomp}
\usepackage{xcolor}
\usepackage{soul}
\usepackage{url}
\usepackage{tabularx}
\usepackage{tcolorbox}
\usepackage{multicol}
\def\BibTeX{{\rm B\kern-.05em{\sc i\kern-.025em b}\kern-.08em
    T\kern-.1667em\lower.7ex\hbox{E}\kern-.125emX}}

\begin{document}

\title{LLM-as-Judge in Education: \\A Curriculum-Grounded Marking Pipeline\\
}

\author{\IEEEauthorblockN{Xiwei Xu}
\IEEEauthorblockA{\textit{CSIRO, Australia} \\
\textit{UNSW, Sydney, Australia}\\
xiwei.xu@csiro.au}
\and
\IEEEauthorblockN{Chen Wang}
\IEEEauthorblockA{\textit{CSIRO, Australia} \\
\textit{UNSW, Sydney, Australia}\\
chen.wang@csiro.au}
\and
\IEEEauthorblockN{Jacky Jiang}
\IEEEauthorblockA{\textit{CSIRO, Australia} \\
jacky.jiang@csiro.au}
\and
\IEEEauthorblockN{Phil Yang}
\IEEEauthorblockA{\textit{Studitory, Australia} \\
studitory@gmail.com}
\and
\IEEEauthorblockN{Qian Fu}
\IEEEauthorblockA{\textit{CSIRO, Australia} \\
qian.fu@csiro.au}
\and
\IEEEauthorblockN{Mohan Dhall }
\IEEEauthorblockA{\textit{Australian Tutoring Association, Australia} \\
mdhall@ata.edu.au}
\and
\IEEEauthorblockN{Wenjie Zhang}
\IEEEauthorblockA{\textit{UNSW, Australia} \\
wenjie.zhang@unsw.edu.au}
\and
\IEEEauthorblockN{Liming Zhu}
\IEEEauthorblockA{\textit{CSIRO, Australia} \\
\textit{UNSW, Sydney, Australia}\\
liming.zhu@csiro.au}
}

\maketitle

\begin{abstract}

Generative AI and large language models (LLMs) are increasingly applied to question generation and automated assessment. However, deploying LLMs in 
preparation for high-stakes exams requires more than prompt engineering; it demands software pipelines that systematically ground model outputs in authorised curriculum artefacts and marking guidelines issued by education authorities. This paper presents a curriculum-grounded, configurable LLM-as-Judge pipeline for question-level marking, co-developed with an industrial partner, to support exam preparation for university admission. The pipeline identifies the relevant topics, subtopics, and cognitive demand of a question, 
and assembles verifiable and authorised context to support LLM judgement. Curriculum intent is operationalised through concrete syllabus artefacts, including prescribed verbs and outcomes, performance band descriptors, glossary definitions, and marking-guideline principles. A staged LLM workflow is employed to first generate question-specific rubrics, capturing structured expectations of performance, and then derive and evaluate marking criteria used to allocate marks to student responses. This design improves consistency, transparency, and alignment with official marking practices. Preliminary evaluation shows that the proposed LLM-as-Judge pipeline delivers marking outcomes comparable to human tutors, while yielding justifications that are more traceable to authorised curriculum artefacts and marking standards. 
The pipeline has also been integrated into an online study platform, where early deployment data provide initial insights into operational usage and manual overrides.
\end{abstract}


\section{Introduction}

Large language models (LLMs) have shown positive influence on learning performance~\cite{wang2025effect} and are increasingly being adopted across the education sector worldwide. Recent industry evidence indicates that AI usage in education has surged, with 86\% of education organisations now using generative AI
\cite{microsoftAIeducation2025, google_our_life_with_ai_2026}. In 
Australia, several state education departments have deployed large-scale educational chatbots that are accessible to thousands of public-school teachers and students\footnote{EduChat from NSW Department of Education (\url{https://education.nsw.gov.au/teaching-and-learning/education-for-a-changing-world/nsweduchat})}\footnote{EdChat from South Australian Department for Education (\url{https://www.education.sa.gov.au/parents-and-families/curriculum-and-learning/ai/edchat})}. In the private sector, conventional tutoring centres have likewise begun exploring LLM-based tools to provide writing feedback and to scale tutoring services through online study platforms, particularly as high-stakes examinations increasingly move to digital delivery\footnote{ OC Selective exam preparation platform (\url{https://ocselective.com.au}
)}\footnote{Studitory: HSC preparation platform (\url{https://studitory.com}
)}.

Among the many educational activities affected by this rapid adoption, assessment lies at the heart of education because it determines how student learning is measured, interpreted, and recognised~\cite{bennett2011formative}. 
At the same time, assessment remains one of the most labour-intensive, subjective, and operationally challenging components of education systems, particularly at scale. Even when detailed rubrics are provided, studies consistently report substantial inter-marker and intra-marker variability, with different assessors, and sometimes the same assessor at different times, assigning divergent scores to identical student work~\cite{Consistency}. One contributing factor to intra-marker variability is cognitive bias: human markers may be influenced by their prior judgements of performance, a phenomenon commonly referred to as the \textit{halo effect}\footnote{\url{https://en.wikipedia.org/wiki/Halo_effect}}.

In high-stakes examinations, such as national assessments~\cite{acara2024naplan}, state level selective school placement examinations\footnote{\url{https://education.nsw.gov.au/schooling/parents-and-carers/choosing-a-school-setting/selective-high-schools/placement-test}}, and university entrance exams, 
hundreds or even thousands of students may fall within a single score band. In such contexts, 
distinguishing students with subtle differences in capability, while providing meaningful and actionable feedback during exam preparation, 
becomes increasingly difficult.

Recent advances in LLMs offer new opportunities to rethink assessment workflows. LLMs have demonstrated strong capabilities in interpreting complex rubrics, generating structured feedback, and reasoning over large volumes of textual data~\cite{LLMGen2024}. 
These capabilities have motivated growing interest in applying generative AI to automated judgment, feedback generation, and adaptive assessment\cite{wang2025effect, gu2025surveyllmasajudge}. 
Conventional assessment workflows are anchored in authorised marking criteria and rubrics that are interpreted and applied by human teachers. Introducing LLMs through basic prompt-based automation risks undermining these properties, as model outputs may drift from authorised standards, introduce inconsistency, or lack traceability to official curriculum sources~\cite{essayScore2021,weidinger2021ethicalsocialrisksharm}. 

A further limitation of current LLM-based approaches lies in their limited capacity to capture \textit{tacit human knowledge}, in addition to the knowledge embedded within LLMs. Experienced teachers rely on implicit judgments, such as coherence of argumentation, density of disciplinary language, and overall quality of reasoning, before formal marking criteria are even applied. This tacit knowledge evolves over time with curriculum updates, cohort characteristics, and disciplinary norms~\cite{polanyi1966tacit}. 

This paper proposes an architecturally operationalised LLM-as-judge pipeline: a curriculum-grounded, verifiable LLM-based system designed for high-stakes educational assessment in exam-preparation contexts. The pipeline is co-developed with an industrial partner and systematically incorporates authorised syllabus documents, performance band descriptors, glossary definitions, and marking guideline principles as structured context for LLM reasoning. The pipeline is deployed in a live online study platform, allowing us to observe the operational signals, like override rates, and robustness against adversarial inputs. 
This paper makes the following contributions:

\begin{itemize}
    \item We designed and implemented a curriculum-grounded, configurable LLM-as-judge pipeline that embeds authorised assessment artefacts as structured context; 
    \item We identified tacit human judgement as a critical but under-represented component of trustworthy educational assessment, and analyse its implications for the design of 
    the system;
    \item We conducted a preliminary evaluation showing that the proposed pipeline achieves outcomes comparable to human tutors, while providing feedback that is explicitly traceable to curriculum intent.
\end{itemize}

The remainder of this paper is organised as follows. After discussing related work in Section~\ref{sec:relatedwork}, Section~\ref{sec:overview} presents the overall architecture of the curriculum-grounded marking pipeline. Sections~\ref{sec:syllabus}–\ref{sec:feedback} describe the design, implementation, and integration of the pipeline within an online study platform. Section~\ref{sec:evaluation} reports preliminary evaluation results from real-world use. Section~\ref{sec:future} discusses future work, and Section~\ref{sec:conclusion} concludes the paper.

\section{Related Work}
\label{sec:relatedwork}

\subsection{LLM-Enabled Question Generation and Assessment} 

Automated assessment has a long research history, particularly in essay scoring and short-answer grading. Early studies evaluated machine-generated scores against human raters, focusing on reliability, validity, and alignment with expert judgement~\cite{ lockwood2014handbook, ramesh2022automated}. These foundational works established evaluation metrics and frameworks for automated scoring systems.

Building on this foundation, the emergence of LLMs has enabled more flexible and powerful assessment methods. Recent work has explored rubric-based and multidimensional evaluation frameworks, demonstrating structured approaches to align LLM-generated scores with human-like assessment criteria~\cite{hashemi2024llm}. 
In parallel, large-scale benchmark studies have explored LLMs as general-purpose evaluators in open-ended settings, showing that they can achieve high alignment with human judgments on tasks such as multi‑turn conversational evaluation~\cite{zheng2023judging}.
To further improve alignment, other work has proposed self-adaptive rubric paradigms to tailor evaluation criteria to individual items~\cite{fan2024sedareval}. But these studies are largely in general benchmark settings rather than curriculum-grounded educational assessment.

In educational contexts, LLMs have been applied to assessment tasks such as question design, automated marking, and feedback generation across both public and private settings~\cite{saEduAI2025}. Prior studies show that LLMs can interpret rubrics, generate structured feedback, and score short-answer or essay responses with consistency comparable to human inter-rater reliability~\cite{AES2023, LLMGrading2025, essayScore2021, Consistency}. 
Also, empirical evidence from national-scale school‑leaving exams shows that LLM‑based assessment can achieve performance comparable to human raters when official curriculum‑based rubrics are operationalised, while also highlighting practical considerations such as bias, rubric implementation, and the need for human oversight in high‑stakes settings~\cite{karjus2026machine}.
However, performance varies across subjects~\cite{wang2025effect}, reflecting biases in training data and leading to inconsistent outcomes. In practice, teacher-developed agentic systems (e.g., Cogniti\footnote{\url{https://cognitoedu.org/}}
) increasingly automate assessment-related tasks to reduce workload and improve scalability.

A closely related body of work focuses on automated question generation. Manual curation of questions by educators is increasingly unsustainable, particularly in the context of frequent syllabus changes that demand continual updates \cite{bennett2011formative}. In response, existing studies have explored how LLMs can generate questions aligned with different cognitive levels, often using Bloom’s Taxonomy as a guiding framework \cite{kurdi2019AQG}. More recent work has examined the alignment between AI-generated questions and established cognitive frameworks, highlighting both the potential and limitations of generative approaches in maintaining pedagogical validity \cite{yaacoub2025}.
Iterative frameworks and structured LLM pipelines have improved question quality, using self-critique and correction loops to refine generated items~\cite{yao2025mcqg} and providing scalable generation and evaluation for multiple-choice questions in educational practice~\cite{mucciaccia2025automatic}.

Despite these advances, the literature largely treats assessment and question generation as model-centric, prompt-based tasks. Curriculum alignment, authorised marking standards, and jurisdiction-specific assessment practices are typically assumed or enforced through ad hoc review, offering limited support for verifiable and curriculum-grounded assessment. 

\subsection{Retrieval-Augmented LLM Pipelines}

Retrieval-augmented generation (RAG) pipelines combine language models with external knowledge sources to improve factual accuracy, controllability, and transparency~\cite{DP_RAG,RAG_Yang}. From an architectural perspective, these pipelines are typically decomposed into modular stages for retrieval, reasoning, generation, and verification, enabling separation of concerns. 
Recent studies have begun to examine RAG pipelines explicitly in educational contexts, focusing on how retrieval, orchestration, and validation can be engineered to support teaching and assessment workflows rather than generic knowledge-intensive tasks~\cite{RAGEdu2025}.

Our design focuses on systematically applying RAG pipelines within the educational assessment domain by grounding them in authorised curriculum artefacts and assessment norms. We embed RAG within a curriculum-governed marking pipeline that explicitly encodes syllabus intent, marking principles, and assessment constraints as structured components.

\subsection{Human-in-the-Loop and Tacit Knowledge Management}

Assessment is inherently a socio-technical process that relies on professional judgment and tacit knowledge developed through experience. Classic work on tacit knowledge emphasises that important aspects of expertise cannot be fully articulated or reduced to explicit rules \cite{polanyi1966tacit}. In educational assessment, experienced teachers may apply implicit criteria before or alongside formal marking rubrics. 

Empirical studies have shown substantial inter-marker and intra-marker variability, even when detailed marking criteria are provided, underscoring both the importance and the subjectivity of human judgement in assessment \cite{Consistency}. Human-in-the-loop approaches \cite{RAGHuman} in AI systems are often proposed to retain accountability, but they may also introduce additional sources of subjectivity \cite{Amershi_Cakmak_Knox_Kulesza_2014}. In contrast, AI-based assessment components offer the potential to improve consistency by applying marking criteria uniformly across large cohorts. In practice, humans are frequently positioned as runtime supervisors or final arbiters, rather than as contributors to the design-time encoding of professional knowledge.

Consequently, existing human-in-the-loop assessment systems rarely provide mechanisms for capturing tacit judgement, marking principles, or curriculum intent as first-class.  
This gap motivates our design 
that embed human expertise 
through authorised documents, historical practice, and principle-based constraints, while retaining scalability and consistency at runtime.

\section{Marking Pipeline Overview}
\label{sec:overview}

\begin{figure}
\centerline{\includegraphics[scale=0.4]{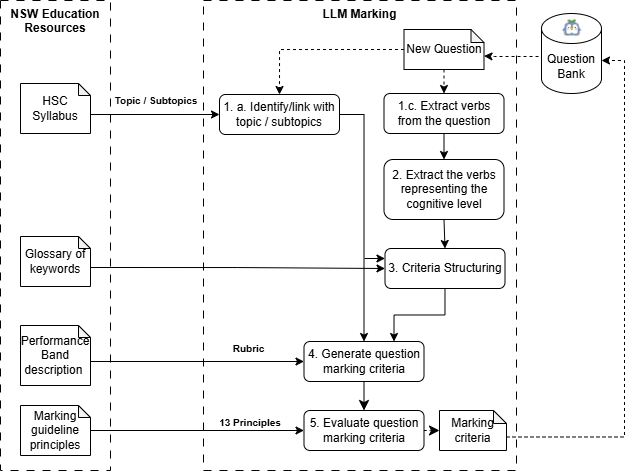}}
\caption{Marking Pipeline.}
\label{fig:markingPipeline}
\end{figure}

Figure~\ref{fig:markingPipeline} illustrates the curriculum-grounded marking pipeline proposed in this paper. The pipeline anchors LLM-assisted marking to authorised artefacts issued by the NSW Department of Education\footnote{\url{https://education.nsw.gov.au/}}, including the HSC (Higher School Certificate) syllabus\footnote{Australia's main secondary school qualification in New South Wales (NSW) for Years 11 and 12, acting as the high school award for university entry}, performance band descriptors, glossary of keywords, and marking guideline principles. These artefacts encode curriculum intent and professional marking practice, and are treated as first-class inputs to the marking pipeline rather than auxiliary context via prompt engineering. By explicitly structuring these sources, the proposed pipeline moves beyond prompt engineering towards a \textit{verifiable} software pipeline that supports consistency, transparency, and curriculum alignment. 

While the core principles of marking are shared across contexts, the operational requirements of official high-stakes examinations and online exam-preparation platforms differ in terms of accountability, timeliness, and tolerance for automation. Official examinations prioritise formal accountability and human authority, whereas online study platforms emphasise scalability, responsiveness, and formative feedback. The marking pipeline presented in this paper is designed for the latter scenario. It was co-developed with our industrial partner, Studitory\footnote{\url{https://www.studitory.app/}}, an online study platform serving over 5,700 registered high-school students at the time of the project, and targets automated marking and feedback for exam-preparation purposes rather than final certification.

The marking process is decomposed into a sequence of verifiable stages. Given a new question either from the existing question bank or generated dynamically, the pipeline first identifies and links relevant syllabus topics and subtopics (\textit{learn about} statements), as well as skills (\textit{learn to} statement) that indicates the cognitive demand of the question, aligning with Bloom's taxonomy. 

The cognitive demand of the question is then inferred by extracting key directive verbs and mapping them to \textit{Glossary of key words} from the NSW Education Standards Authority (NESA), enabling the selection of verbs representing the same cognitive level. The glossary of key words also include NESA-specific marking expectations (e.g. \textit{Explain} MUST use cause-effect pairs). These signals are combined with performance band descriptors, to generate a question-specific marking criteria. The marking criteria provides a structured representation of 1) behaviour-level criteria (Table~\ref{tab:analytic_criteria_example}) and 2) holistic mark-band descriptors (Table~\ref{tab:example_marking_criteria}). 

To support trustworthy use in assessment settings, the pipeline incorporates multiple verification points that evaluate intermediate artefacts against authorised curriculum documents and pedagogical foundations. For example, when skills, concepts, and outcomes are matched, a dedicated verification step checks whether the expected outcomes are appropriately covered by the selected skills and concepts, thereby meeting the overall assessment design goal. Similarly, the marking criteria are generated against recognised marking-guideline principles. The intermediate artifacts are inspectable and persisted for audit.  

The verification function as a governance mechanism that constrains LLM outputs within authorised assessment norms. While the pipeline does not require continuous human intervention at runtime, it is grounded in human expertise embedded at design time through curriculum documents, and institutional and normative constraints that embed historical marking practices. Our design focuses on systematically applying RAG pipelines within the educational assessment domain by grounding them in authorised curriculum artefacts and assessment norms. 


\section{Syllabus Matching}
\label{sec:syllabus}

\subsection{HSC Syllabus}

As shown in Figure~\ref{fig:markingPipeline}, the first key piece of authorized information from the NSW Education is HSC Syllabus, which provides the grounding basis for learning and preparing the exam, and the assessment. These Syllabus artefacts are treated as structured and queryable system components rather than static reference text.

\textbf{Learn about} statement defines the conceptual and content knowledge that students are expected to understand within a topic. They specify the key ideas, models, facts, structures, and subject-specific concepts.

\textbf{Learn to} statement articulates the learning actions and cognitive skills that students are expected to practise using the prescribed conceptual and content knowledge. They are typically expressed through directive verbs (e.g., analyse, explain, evaluate) that align with \textit{Bloom Taxonomy}~\cite{bloom1956taxonomy,anderson2001taxonomy} and describe how students should engage with, manipulate, and apply conceptual knowledge during learning activities. Bloom’s Taxonomy is a hierarchical framework that classifies learning objectives as a ladder of cognitive complexity, from basic knowledge recall to higher-order thinking such as analysis, evaluation, and creation. It provides a foundational pedagogical basis for the design and operation of the marking pipeline.

\textbf{Outcomes} specify the assessable capabilities that students are expected to demonstrate by the end of a stage of learning. Outcomes integrate content knowledge and cognitive skills into higher-level performance expectations and serve as the primary reference for design of marking criteria.

\subsection{Similarity Matching}

The three statements, including \textit{learn to}, \textit{learn about}, and \textit{outcome}, are extracted from authorised syllabus documents provided by the education department. The relevant content is extracted and persisted in a structured database to support downstream querying. A vector database is constructed to store embeddings of these statements, enabling semantic similarity assessment against a given question. The selected \textit{learn to} and \textit{learn about} statements are ranked according to their similarity scores against the given question. \textit{Outcome} statements are broad, and determined by top-ranked \textit{learn to} and \textit{learn about}.


To further select the relevant skill set from the ranked \textit{learn to} statements, the maximum mark allocated to the question ($mark_{max}$) is used as a proxy for question scope and complexity. Specifically, $mark_{max}$ provides an indication of how many distinct skills are expected to be selected and assessed. Based on an analysis of past examination papers over the last seven years (public accessible), we derive a configurable mapping between $mark_{max}$ and the number of skills to be selected. This mapping is designed to be adaptable to different examination formats and assessment designs. Semantic similarity is used to narrow the candidate space, but curriculum coherence is enforced through the subsequent verification stage.

\begin{table}[t]
\centering
\caption{Number of Selected Skills According to $mark_{max}$}
\label{tab:skill_selection}
\begin{tabular}{lc}
\hline
\textbf{No. Top Skills} & \textbf{$mark_{max}$} \\
\hline
Top 1 & 1--5 \\
Top 2 & 6--10 \\
Top 3 & 11--15 \\
Top 4 & 16--20 \\
\hline
\end{tabular}
\end{table}

\subsection{Coherence Verification}

One way to verify the selected skills, concepts, and outcomes is to assess their internal consistency and coherence through cross-checking the coverage of outcomes. This is achieved by measuring the extent to which the selected skill and concept set covers the expected outcome set. We define outcome recall as:

\[
\frac{\#\,\textit{overlaps with expected outcomes}}{\textit{expected outcomes}} \times 100\%
\]

By the end of this stage, each question is linked to a verified set of skills, concepts, and outcomes for assessment. Candidate concept sets that fail to meet outcome coverage thresholds are either re-ranked or discarded, preventing incoherent curriculum mappings from propagating to downstream stages of the pipeline.

\section{Marking Criteria Generation}
\label{sec:markingcriteria}

The marking-criteria generation stage translates curriculum-grounded intent into question-specific, assessable, and verifiable marking artefacts. As illustrated in Figure~\ref{fig:markingPipeline}, this stage integrates a couple of more authorised sources, including glossary of keywords definition, performance band descriptors, marking guideline principles. 
The marking criteria are generated through a two-stage process that separates marking criteria structuring from performance calibration, mirroring established human marking practice.

\subsection{Criteria Structuring}

Following syllabus matching and verification, this stage clusters the matched syllabus skills, concepts, and outcomes into a number of functional assessment dimensions. Each dimension captures a distinct aspect of what the question assesses (e.g., conceptual understanding, application, or reasoning). These dimensions are formalised as criteria, which specify what evidence is being assessed, independently of overall performance quality.

The \textit{HSC Keywords Glossary} is incorporated as a semantic constraint in the marking-criteria generation stage. 
It provides authorised, jurisdiction-specific interpretations of directive verbs that appear in both the question text and the marking criteria. During generation, glossary definitions are injected to stabilise the meaning of key terms (e.g., analyse, account, evaluate), ensuring that marking criteria reflect official expectations rather than generic LLM interpretations. This prevents over-specification or misalignment in feedback language and supports consistent recognition of equivalent merit across alternative correct responses. 

Another primary input to the criteria structuring stage is the maximum mark allocation ($mark_{max}$). The value of $mark_{max}$ plays a critical calibration role: it constrains the granularity of criteria, the number of assessable elements, and the distribution of marks across these elements. Automatically derived match counts, such as the number of relevant skills or concepts identified earlier in the pipeline, are scaled and normalised against $mark_{max}$ to ensure alignment with established assessment expectations. Table~\ref{tab:analytic_criteria_example} illustrates an example structured criteria for an economics question generated from this stage.

\begin{table*}[h]
\renewcommand{\arraystretch}{1.3}
\caption{Example Structured Criteria Generated in the Criteria Structuring Stage}
\label{tab:analytic_criteria_example}
\centering
\begin{tabularx}{\textwidth}{|p{1cm}|p{3cm}|p{3cm}|X|}
\hline
\textbf{Outcome} & \textbf{Skill} & \textbf{Concept} & \textbf{Expected Behaviour} \\
\hline
H4 &
Assess the impact of recent changes in the global economy on Australia’s trade and financial flows &
Implications for Australia of protectionist policies of other countries and trading blocs &
Explains clear cause--effect links between specific protectionist measures imposed by other countries (e.g.\ tariffs or quotas on Australian exports) and changes in relevant components of Australia’s current account (e.g.\ export credits, import debits, net primary income), clearly relating how and why these flows are affected. \\
\hline
H6 &
Discuss the effects of protectionist policies on the global economy &
Australia’s Balance of Payments, including links between key Balance of Payments categories &
Explains how protectionist policies abroad alter the structure and interrelationships within Australia’s current account (e.g.\ deterioration in the goods and services balance, shifts in income flows associated with foreign investment), making the links between protection, trade flows, and current account outcomes explicit through coherent cause--effect reasoning. \\
\hline
\end{tabularx}
\end{table*}

\subsection{Performance Calibration}

The second stage translates structured criteria into quality levels that describe \textit{how well} each criterion is demonstrated. This stage operationalises authorised marking standards by aligning criterion-level expectations with \textit{Performance Band Descriptions}. Given the question context, including the $mark_{max}$, directive verb, and the set of structured criteria produced in the previous stage, the second stage synthesises each criterion into four conceptual quality levels: \textit{Comprehensive}, \textit{Sound}, \textit{Some Understanding}, and \textit{Limited}. The descriptors associated with these levels are explicitly constrained and styled using authorised performance band descriptions, ensuring alignment with official marking practice rather than generic model interpretations.

The granularity of performance calibration is dynamically adjusted based on $mark_{max}$. For higher-mark questions (e.g., $mark_{max}\geq$15 marks), the \textit{Some Understanding} level is subdivided to support finer distinction between partially correct responses. For low-mark questions (e.g., $mark_{max}\leq$5 marks), quality levels are compressed to reflect the reduced resolution typically expected in official marking schemes.

Table~\ref{tab:example_marking_criteria} illustrates a representative performance calibration for the same economics question, where behaviour-level criteria specify the evidence required for full and partial credit. The criteria are explicitly aligned with both conceptual understanding and procedural execution.

\begin{table*}[h]
\renewcommand{\arraystretch}{1.3}
\caption{Example Marking Criteria of a Economics Question Generated In The Performance Calibration Stage ($mark_{max}: 5$ )}
\label{tab:example_marking_criteria}
\centering
\begin{tabularx}{\textwidth}{|X|c|}
\hline
\textbf{Criteria} & \textbf{Marks} \\
\hline
Explains clear cause–effect links between specific protectionist policies imposed by other countries (e.g. a tariff or quota on Australian iron ore, coal, tourism, education services) and changes to identified components of Australia’s current account. Correctly relates how and why these measures reduce export credits and/or alter import debits, and may extend to effects on net primary income (e.g. lower profits/dividends on foreign‑owned firms in Australia due to reduced export sales). Makes the relationships between foreign protection, trade flows and the goods and services balance/current account balance evident using appropriate economic terminology and at least one well-explained cause–effect pair (often more). & 4-5 \\
\hline
Describes some relevant protectionist policies by other countries and shows partial understanding of their impact on Australia’s current account, but cause–effect links are underdeveloped, incomplete or only partly accurate. May correctly identify a cause (e.g. tariff on Australian exports) and a related current account component (e.g. exports of goods) but explains the effect briefly or in a general way (e.g. ``worsens the current account'') without clearly linking how and why export credits, import debits or net primary income change. Uses some economic terms but with limited detail or precision. & 2-3 \\
\hline
Provides a very limited or inaccurate description of protectionist policies and/or Australia’s current account, with little or no valid cause–effect explanation. May make vague statements about `less trade'' or ``hurting the economy'' without linking to specific current account components, or shows serious misconceptions about the nature of the current account or protectionism. Responses that are off-topic, purely definitional, or too brief to establish a causal relationship fall in this range. & 1 \\
\hline
\end{tabularx}
\end{table*}





\subsection{Normative Constraints Verification}

To provide stable reference patterns, 
and reduce hallucinated assessment norms, the pipeline explicitly incorporates \textit{marking guideline principles} as a normative constraint. These thirteen principles, such as consistency, fairness, allowance for alternative correct approaches, and proportional allocation of marks, are retrieved from official marking guideline documents and embedded into the prompt design as verified rules that impose non-negotiable boundaries on LLM behaviour.


By integrating these principles during generation, the pipeline discourages over-specification of model answers, enforces explicit support for partial credit and equivalent merit, and aligns mark allocation with established professional practice. In this way, adherence to authorised marking norms is achieved through constrained generation, rather than post-hoc filtering or correction. At the end of this stage, the generated and verified marking criteria are persistently linked to the question and stored for subsequent assessment when the question is selected by a student. (Appendix~\ref{app:marking_guideline} summarises the alignment between the proposed marking pipeline and the thirteen marking guideline principles.)

\section{Marking and Feedback Loop}
\label{sec:feedback}

With the marking criteria associated with given question, this stage closes the loop between syllabus-grounded design-time artefacts and runtime assessment behaviour, ensuring that marking remains consistent, 
and enabling continuous improvement.

\subsection{Automated Marking}


Once a student submits a response, the attached marking criteria and guidelines are retrieved and used to score the response on a per-criterion basis. Each criterion is evaluated independently, guided by the expected evidence cues and performance bands defined during criteria generation. Partial credit is supported by being incorporated in the marking criteria (Table~\ref{tab:example_marking_criteria}).

\subsection{Skill-Aligned Feedback Generation}

In parallel with scoring, the marking pipeline generates an overall mark with justification referring to the marking criteria, which is explicitly aligned with the syllabus skills identified earlier in the pipeline. Rather than producing generic comments, feedback is grounded in the same syllabus anchors used for marking, enabling students to understand not only what was incorrect or incomplete, but which outcome or skill was not sufficiently demonstrated and at what cognitive level.

This alignment ensures coherence between assessment, feedback, and learning objectives, and supports downstream uses such as targeted remediation, and adaptive learning. This design mitigates misalignment between automated feedback and assessment criteria, a challenge in LLM-based tutoring and assessment systems~\cite{bennett2011formative, AES2023}.


\subsection{Process and Provenance Verification}

The marking stage incorporates \textit{process and provenance verification} to support trustworthiness without introducing additional runtime human judgement or intervention. This form of verification focuses on ensuring that the assessment process itself remains transparent, inspectable, and reconstructable. All intermediate artefacts produced during the process are persistently recorded with timestamps and contextual metadata. This enables each marking decision and feedback instance to be traced back to the specific structured criteria, performance band constraints, and calibration logic that shaped its generation. By preserving provenance across the assessment lifecycle, the pipeline allows marking behaviour to be audited, reproduced, and examined post hoc across different runs and model configurations. 

\section{Architectural and Design Discussion}

This section reflects on two broader design implications of the proposed marking pipeline: \textit{pipeline dynamism} as a mechanism for sustainable evolution under changing curricula and assessment norms, and the operationalisation of \textit{tacit professional judgement} through structured artefacts and constraints. Together, these implications highlight how trustworthiness emerges from architectural design choices rather than from isolated model behaviours.

\subsection{Embedding Implicit Professional Knowledge}

The professional knowledge applied during live marking is not purely tacit, some has already been articulated as holistic qualities within authorised assessment artefacts, such as performance band descriptors and generic rubrics. In practice, human teachers apply these standards rapidly and informally, drawing on experience to interpret and operationalise them during marking. 

There are roughly three levels of implicit professional judgement in assessment. First, codified authorised practice includes artefacts such as performance band descriptors, glossary-defined directive verbs, and marking guideline principles. These forms of knowledge are not tacit, as they are formally specified, but are often applied heuristically by human markers. Second, marker heuristics encompass experience-based practices, including awareness of common misconceptions, judgements about what constitutes ``sufficient'' evidence cue, and typical partial-credit allocations; these are closer to tacit knowledge and are shaped through repeated marking experience. Third, moderation and standard-setting operate at an institutional level, involving cohort calibration and the evolution of standards over time; this form of judgement is procedural and collective, and is not captured by default by LLMs. This norm-referencing against peers conflates outcomes assessment with peer-referenced approaches.

The proposed marking pipeline primarily supports the first level by encoding authorised, codified assessment authorized documents, including syllabus outcomes and content, directive verb semantics from the NESA glossary, performance band descriptors, and marking guideline principles, into explicit, machine-verifiable rubric artefacts. It partially supports the second level of marker heuristics by embedding conservative, criteria-first decision rules (e.g., prioritising behaviour-level analytic criteria and using holistic bands as a consistency check), and by promoting fine-grained error detection through structured justifications. While the pipeline does not automate moderation or standard-setting, it produces configurable, moderation-ready artefacts, such as traceable matching results, question-specific guidelines, and criterion-referenced marking decisions, which enable human review, calibration, and iterative refinement within established governance processes.

\subsection{Pipeline Dynamism}

Pipeline dynamism refers to the ability of the marking pipeline to adapt across pedagogical contexts and evolve over time. In the proposed design, dynamism is achieved through a combination of explicit design-time configuration and evidence-driven system evolution. Design-time configuration enables controlled variation in how syllabus signals, skills, and concepts are selected and weighted, while runtime telemetry supports continuous refinement of these configurations based on observed assessment behaviour. Together, these mechanisms allow the pipeline to remain flexible and responsive without compromising verifiability, or curriculum alignment. 

\subsubsection{Configurable Retrieval and Ranking Strategies}

The marking pipeline exposes several configurable options that allow it to be adapted to different assessment contexts while remaining aligned with pedagogical intent. One such option is the application of post-filtering or re-ranking weights based on outcome coverage when selecting the top-ranked skills. For example, skills that explicitly cover the expected syllabus topics or outcomes inferred from the question text can be prioritized. Similarly, when an expected concept can be inferred with high confidence, it may be explicitly seeded into the concept shortlist or assigned a higher weight during retrieval. 

\subsubsection{Telemetry and Maintainability}

To support systematic refinement, the marking and feedback loop instruments a range of telemetry signals during runtime. These include coverage metrics indicating how well retrieved outcomes and skills are reflected in student responses, rubric confidence scores capturing the strength of alignment between expected evidence cues and observed responses, and marking variance across multiple runs or model configurations where applicable.

These signals are aggregated and analysed offline to inform adjustments to retrieval thresholds, clustering strategies, prompt templates, and rubric construction parameters. 
This telemetry-driven design enables failures to be \textit{localised} and addressed at the appropriate stage of the pipeline. When marking behaviour appears inconsistent or misaligned, issues can be traced back to specific upstream components, such as syllabus matching, criteria structuring, or performance calibration, and corrected by updating the corresponding configurations or generation logic. Revised artefacts can then be regenerated deterministically, without retraining human markers or relying on informal calibration processes. While the pipeline may still fail in cases of incorrect retrieval or poorly generated criteria, its modular structure makes such failures inspectable and correctable. A further factor limiting its effectiveness is where a syllabus, outcomes and other content or skills elements are actually not stated, but are assumed. This can occur when a syllabus references a concept but not the depth or scope of the concept. Under such circumstances extraneous materials, such as published textbooks for courses, can provide interpretive support.

\section{Preliminary Evaluation}
\label{sec:evaluation}

The goal of the preliminary evaluation is not to measure learning outcomes, but to assess whether the proposed architecture exhibits the system-level properties required for trustworthy use in curriculum-governed assessment. We evaluated both the assigned marks and the generated justifications, focusing on their consistency and traceability to authorised curriculum artefacts.

\subsection{Dataset and Metrics}

The marking dataset contains 47 questions with marks given by a single tutor. The tutor's marks are not perfect and contain few errors due to time pressure of marking. However, these marks are considered acceptable and representative of marking practices of tutors under time pressure. We use these marks as an acceptable baseline to evaluate the agreement of marks produced by our pipeline. We further use direct LLM prompting as additional baselines in the evaluation (the prompt template is shown in Appendix~\ref{sec:direct_prompt}. Meanwhile, we carefully examine the marking details of a few cases to compare the subtle differences empirically. To do these, we randomly sample 10 questions to form a pool of few-shot examples for LLMs. This is to mimic the calibration process in human marking. 
The remaining 37 questions are used to evaluate the performance of different methods. 

We use Lin's concordance correlation coefficient (CCC) and weighted Cohen's kappa to examine the difference between human marking and LLM marking. 

The CCC, denoted by $\rho_c$ is defined as below:
\[
\rho_c \;=\; \frac{2\rho \sigma_x\sigma_y}{\sigma_x^2+\sigma_y^2+(\mu_x-\mu_y)^2}
\]
in which, $\{(x_i,y_i)\}_{i=1}^n$ is the paired marking of question $i$ between human and the LLM; $\rho$ represents the Pearson correlation of the paired marks; $\mu_x,\mu_y$ are the corresponding mark means and $\sigma_x^2,\sigma_y^2$ are the mark variances. CCC measures how closely paired marks fall on the line of perfect concordance $y=x$. 

The weighted Cohen's kappa provides a robust measurement of inter-marker agreement. It assigns different weights to disagreements based on the magnitude of the difference. The weighted Cohen's kappa is defined as below:

\[
\kappa_w
=
\frac{
\sum_{i=1}^{k}\sum_{j=1}^{k} w_{ij}\,o_{ij}
-
\sum_{i=1}^{k}\sum_{j=1}^{k} w_{ij}\,e_{ij}
}{
1-\sum_{i=1}^{k}\sum_{j=1}^{k} w_{ij}\,e_{ij}
}
\]
Where $w_{ij}$ is the weight associated with the degree of disagreement, $o_{ij}$ is the observed proportion, and $e_{ij}$ is the expected proportion by chance.

We also use BLEU (bilingual evaluation understudy) score to measure the alignment of generated text. BLEU is a metric for evaluating generated text by measuring how much its n-grams overlap with one or more reference texts.

\begin{figure}
\centerline{\includegraphics[scale=0.4]{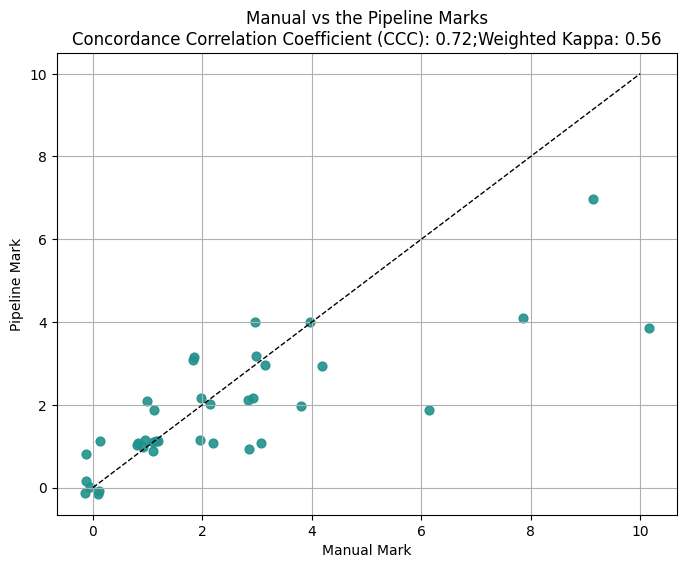}}
\caption{Comparison between human marks and our pipeline produced marks (gpt-5 is used in marking criteria generation and evaluation).}
\label{fig:human_vs_pipeline}
\end{figure}

\begin{figure}
\centerline{\includegraphics[scale=0.4]{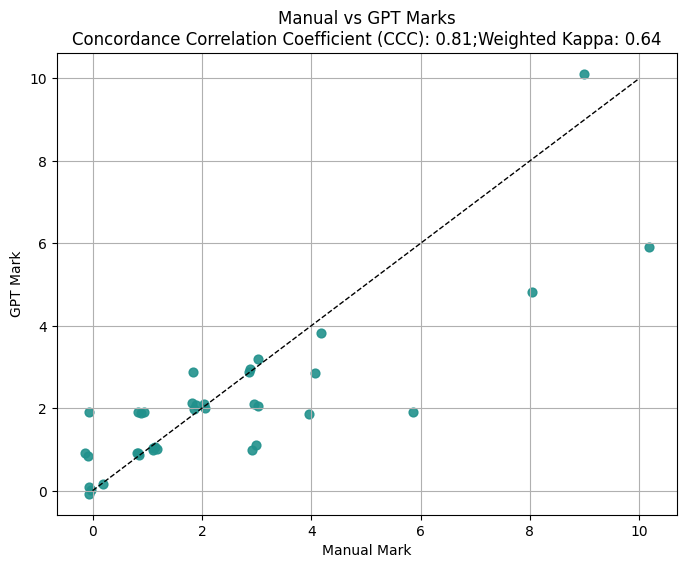}}
\caption{Comparison between human marks and LLM marks: providing syllabus and learning outcome descriptions for LLMs to predict learning outcomes associated with the questions before marking (gpt-5 is used in marking).}
\label{fig:human_vs_gpt5}
\end{figure}

\begin{figure}
\centerline{\includegraphics[scale=0.4]{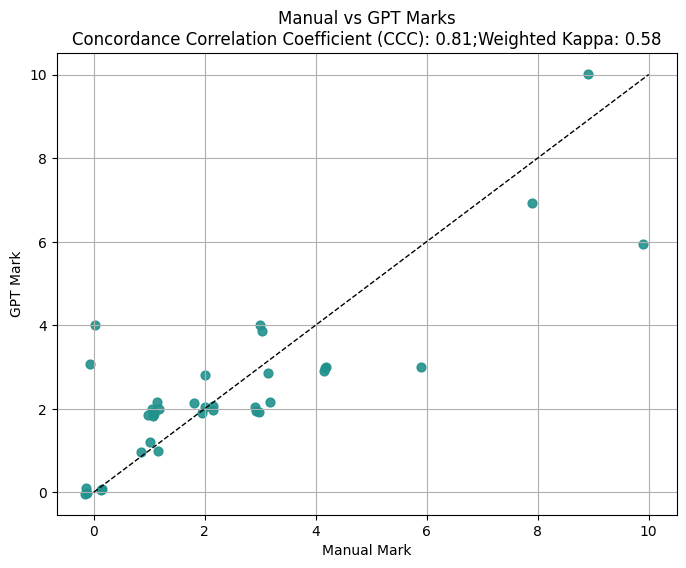}}
\caption{Comparison between human marks and LLM marks: results with a smaller model (gpt-5-nano is used in marking).}
\label{fig:human_vs_gpt5-nano}
\end{figure}

\begin{figure}
\centerline{\includegraphics[scale=0.4]{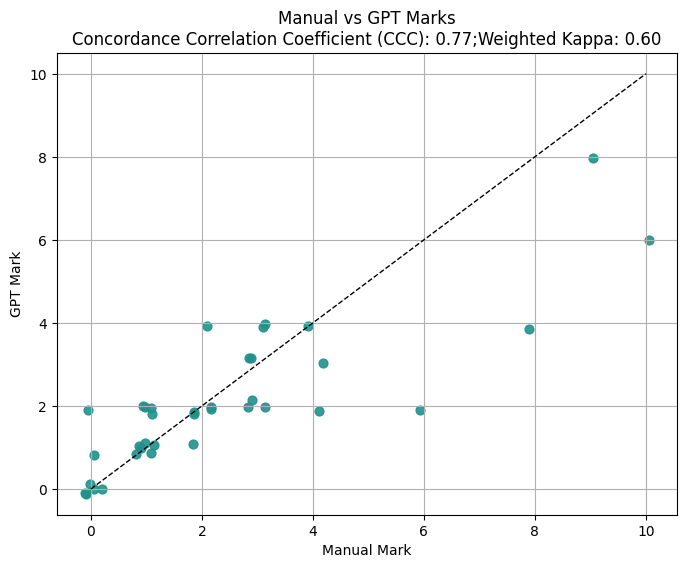}}
\caption{Comparison between human marks and LLM marks: providing random learning outcomes to the questions before marking.}
\label{fig:human_vs_gpt5-nano_no_context}
\end{figure}

\subsection{Human and LLM Agreement on Marks} 

We measure the agreement between the human and LLMs on the dataset using these metrics. We intend to answer the following questions:
\begin{enumerate}
    \item How do the marks produced by direct LLM prompting compare to those by the human?
    \item What is the impact of model size on the marking performance?
\end{enumerate}

Fig.~\ref{fig:human_vs_pipeline} shows the agreement between the marks produced by our pipeline and those given by the human marker. Fig.~\ref{fig:human_vs_gpt5} shows the marking agreement between the LLM (gpt-5) and human with CCC equals to 0.81 and the weighted Cohen's kappa of 0.64. In this experiment, we provide syllabus, verbs and 5 randomly sampled questions together with their marks to prompt the LLM directly to mark each question in the test set and provide justifications. The model is prompted to predict the learning outcome of each question first before marking them. The result indicates that the knowledge in the LLM paired with external data mimic the human marks better than the process-oriented approach of our pipeline.   

To further confirm the knowledge in a model plays a significant role in marking questions in our dataset, we switch the LLM from gpt-5 to gpt-5-nano to check to agreement change. As shown in Fig.~\ref{fig:human_vs_gpt5-nano}, the CCC scores are comparable while the kappa score of the smaller model is lower. This indicates that both models can produce similar marking scores comparable to human. In Fig.~\ref{fig:human_vs_gpt5-nano_no_context}, we examine if providing random learning outcomes sampled from all outcomes in the syllabus affect the marking of the smaller model. The CCC values and kappa scores are different to those in Fig.\ref{fig:human_vs_gpt5-nano}, but the differences may be insignificant considering the small number of questions. 

LLMs can clearly leverage internal knowledge to mimic human marking outcomes numerically. They outperform our pipeline that decomposes the marking to a sequence of processes in mark agreement with human. This may be explained by the effect of RLHF (Reinforcement Learning from Human Feedback) in LLM training. The scalar ratings in human feedbacks make LLMs sensitive to scores rather than score justifications~\cite{rafailov2024scaling}. We therefore hypothesize that while unguided LLMs may approximate human marks numerically, they do not provide guarantees of alignment with authorised curriculum artifacts, nor do they produce justifications that are traceable to official assessment artefacts.

We measure the effect in the next subsection.





\subsection{Comparison of LLM Generated Marking Justifications}

We compare the justification generated by our pipeline with those generated by gpt-5 and gpt-5-nano. We evaluate whether the generated written justification appropriately explains and supports the assigned mark. Specifically, we examine its alignment with the task requirements and directive verbs, the explicit referencing of evidence in the student response, the clarity and specificity of the explanation, and its usefulness in helping students to understand. 


To do this, we conducted a pairwise comparison between the pipeline justifications and  justifications produced by the human tutor and other methods. For each question, the pipeline output is compared independently against each baseline, yielding three pairwise comparisons per question and a total of 37 paired items per comparison. Pairwise comparisons are assessed by two independent LLM judges drawn from different model families than the systems under evaluation, in order to reduce model-specific bias and surface disagreement as uncertainty. The judging criteria are drawn from NSW marking principles\footnote{\url{https://www.nsw.gov.au/education-and-training/nesa/hsc/exams-and-marking/marking-guideline-principles}}. Due to model availability, the initial evaluation uses Claude Opus 4.5 and Qwen3 (32B). For each comparison, outputs are anonymised and randomly ordered as Response A and Response B, with no system identity revealed. Judges are provided with the question text, maximum mark, directive verb definition (where available), student response, and both candidate outputs. Judges return a \textit{win/tie/loss} decision supported by short, evidence-based reasoning. Outcomes are encoded numerically (win = 1.0, tie = 0.5, loss = 0.0) and averaged across judges, yielding a single score in $[0,1]$ per question, per comparison.

Table~\ref{tab:justification_quality} summarises the results of the pairwise evaluation on justification quality. Across all comparisons, the proposed pipeline clearly achieves a higher win-rate than both human and single-call LLM baselines, with the largest margin observed against human-generated justifications.

\begin{table}[t]
\centering
\caption{Justification Quality: Pairwise Comparison Results}
\label{tab:justification_quality}
\begin{tabular}{lcccc}
\hline
\textbf{Comparison} & \textbf{Win-rate} & \textbf{Win} & \textbf{Tie} & \textbf{Loss} \\
\hline
pipeline vs. human     & 80.4\% & 26 & 11 & 0 \\
pipeline vs. gpt-5     & 69.6\% & 19 & 17 & 1 \\
pipeline vs. gpt-5-nano& 69.6\% & 17 & 19 & 1 \\
\hline
\end{tabular}
\end{table}

Table~\ref{tab:bleu} further shows the BLEU scores of those justification against wording in syllabus, glossary of key words in NSW HSC marking guidelines and performance band descriptions. Clearly the pipeline generates justification overlapping better in wording with the text in provided materials than human justifications and justifications produced by direct prompting (The prompt of assessment judge is shown in Appendix~\ref{app:judge}). 

\begin{table}[ht]
\centering
\caption{Comparison of marking justifications of different LLM marking methods.}
\begin{tabular}{l r}
\hline
Marking method & BLEU score \\
\hline
pipeline & 8.457e-03 \\
gpt-5 & 0.191e-03 \\
gpt-5-nano & 0.491e-03 \\
human & 0.080e-03 \\
\hline
\end{tabular}
\label{tab:bleu}
\end{table}

\textbf{Discussions.} An LLM may already contain sufficient general pedagogical knowledge to perform marking without explicitly relying on provided learning outcomes, as it can implicitly capture normative assessment patterns learned from pedagogical theories and examples, similar to those internalised by human tutors. This, plus the effect of scalar rewarding signals in LLM training, can explain why the unguided LLM achieves scores comparable to those of human markers. However, when assessment must align with country- or jurisdiction-specific curricula and provide practical, exam-oriented feedback to students, explicit grounding in the authorised curriculum is critical. On the other hand, we cannot simply conclude that the proposed pipeline justification can replace human justification as the BLEU scores can also be seen from a novelty angle where human justification, or direct LLM justification may provide creative comments that are particularly useful to students.

\subsection{Case study}

Two illustrative case studies are discussed to highlight instances where the human marker and the AI system assign different marks. Rather than treating these disagreements as errors, the cases are used to examine how subtle differences in judgement arise in practice. Specifically, we analyse Question \#13936 and Question \#13966 to compare the marking rationales applied by human markers and the proposed pipeline.

\subsubsection{Question \#13966}

This question asks students to \textit{Explain ONE positive impact and ONE negative impact of transnational corporations (TNCs) on developing countries (4 marks)}. The student response contained severe spelling and grammatical errors: \textit{good they allow for icnreae laboru migriaon and for ceap producion csots … bad – they cna avoid taxe and degrtae hte nvroennt}. The human marker awarded 0 marks, accompanied by the comment: \textit{Incomprehensible. As an AI, you may be able to decipher this. However, this should not be the responsibility of the teacher.} 


This mark reflects a conservative approach to marking where the marker judges the grammar rather than the content and intention of the student. The critique is personal rather than focused on the student’s understanding. The initial impression of written work to a marker is important when shaping a response. Marker’s typically judge first whether a student is high, middle or low achiever.  They second step is according a mark within a band range that aligns to the marker’s perception of the response. The marker chose not to invest effort in reconstructing the intended meaning of a response that falls below an acceptable readability threshold, and therefore treated the answer as non-responsive. It also demonstrates a negative marking approach - inconsistent with contemporary approaches to marking which favour markers looking for what is correct rather than penalising on the basis of what is deemed to be a response weakness.

In contrast, the AI marking pipeline awarded 1 mark and provided an explicit justification. The pipeline attempted to recover partial meaning from the response, identifying fragments that could plausibly be interpreted as a negative impact of TNCs, namely tax avoidance and environmental degradation. However, it explicitly noted that neither the positive nor the negative impact was explained through a clear cause–effect link to economic outcomes in developing countries, as required by the marking criteria. The AI therefore placed the response in the 0–1 band.

This case highlights a key source of disagreement: the AI applies a criteria-first interpretation that is tolerant of partial semantic recovery, whereas the human marker exercised a pragmatic legibility cutoff, opting not to reconstruct meaning when the response quality fell below an informal threshold. It also highlights how the AI did not personalise the comments or appeal to the student to make the work resonate with the needs of marker. This aspect is significant and needs further understanding.

\subsubsection{Question \#13936}

The question is to \textit{Discuss the consequences of a sustained worsening of Australia’s Current Account Deficit.} It carries a maximum of 4 marks and uses the verb \textit{discuss}, which, according to authorised glossary guidance, \textit{can mention both sides of the argument with each side should be marked independently, OR identify issues and provide points for and/or against}. The pipeline was set to take the approach that answer MUST refer to arguments for/AND against rather than points for/OR against.


The student response demonstrates a solid understanding of negative consequences, but does not address any mitigating effects. This is reflected in the criterion-level reasoning: \textit{The response demonstrates a reasonable understanding of some adverse consequences of a sustained worsening CAD, but does not address any offsetting/mitigating consequences, so it does not meet the ‘discuss both sides’ requirement.}



Because the pipeline explicitly disallows compensating for a missing side of the argument, the final mark is constrained to the lowest band: \textit{As only one side of the argument is presented, the response falls into the 0–1 band, but given the solid treatment of the negative side, a mark of 1 is appropriate rather than 0.} 


The human marker awarded 4 marks, describing the response as a \textit{model response} that clearly outlines consequences and explains their effects at a detailed level. This case illustrates how disagreements between human and AI marking can arise from different interpretations of task fulfilment based on an interpretation of a directive term. By enforcing a fixed though incorrect approach to the semantics of directive verbs, the AI system applies marking rules consistently, even when this leads to conservative or incorrect outcomes. It should be noted however that there has been some teaching professional-level discussion about the use of the term ``Discuss" and whether both sides of an argument must be mentioned or not. In this case, normative practice needs to guide the pipeline inputs. Note also that without being able to see the student’s response in full we cannot assess whether 4 marks is also warranted. This is because in the recent past Australia’s foreign debt has been mostly characterised in AUD not foreign currencies – a ppoint overlooked by most economics teachers who teach from a historical rather than contemporary perspective.

\section{Deployment in the Studitory}
\label{deployment}

The proposed marking pipeline has been integrated into the Studitory study platform, serving over 5,700 registered students at the time of the project. The deployed pipeline was used to generate marks and feedback for student responses submitted through the platform.

A feedback report from platform logs provides an initial view of real-world usage. Between 31 January and 7 March 2026, the system recorded 544 AI marking log entries, while 3,166 answers were processed between 31 January and 7 March 2026 after the marking engine went live. During this post-launch period, only 92 answers were manually overridden, corresponding to an override rate of 2.91\%. These data suggest that the deployed pipeline was operationally usable in the platform and that most marks were accepted without manual correction, although override logs alone do not reveal whether users agreed with the marks or simply did not revise them. Although the platform includes fields for user feedback ratings, comments, and perceived scores, none of these fields had yet been populated, and no manual reasons were recorded for overrides. This limits our ability to interpret why disagreements occur and points to an important next step for improving the evaluation of real-world educational AI systems. Recent logs included prompt-injection attempts, and the system assigned 0 marks in all observed cases, providing early evidence that the deployed workflow retained robustness under adversarial student inputs.

\section{Discussion and Research Agenda}
\label{sec:future}

The AI marking has significant implications for student assessment. With iterations made to the marking pipeline to be inclusive of nuance informed by professional teaching practice it would be expected to mark and give feedback on short answer questions with a high degree of accuracy. Moreover, the feedback can be linked to the identifiable elements of achievement thereby providing a basis to guide student growth. This is important in assisting student understand how to use feedback for academic growth. The AI marking does not make subjective judgements about the writer. This is significant as it can mark free of history or a student’s personality. In schools where student names are routinely used, instead of student numbers, the use of AI can guide professional judgement about the work, rather than the person. A further area of application could be in Initial Teacher Education (ITE) programs. Typically, methods lecturers focus on teaching strategies rather than assessment practice, and leave assessment practice to the profession. This makes the young in career teacher vulnerable, especially when teaching a senior subject for the first time.

This work opens a broader research agenda at the intersection of software architecture, educational assessment, and responsible generative AI. 

\subsubsection{Extending from question marking to question generation}

The current pipeline focuses on question-level marking and feedback. Future work will extend this architecture to support question generation, enabling the systematic expansion of the industrial partner’s question bank. Significant portions of the existing pipeline can be reused, including syllabus matching, outcome alignment, and cognitive-level inference. However, question generation introduces additional challenges for adaptive learning, such as ensuring cumulative skill coverage, controlling progression of cognitive demand across generated questions, and avoiding redundancy or unintended curriculum gaps. Including into the pipeline, content, knowledge or skills not made explicit in curriculum but relevant to the examinable aspects of subject areas can also be considered.



\subsubsection{Generalisation across jurisdictions and subjects}

Future research will explore how the proposed pipeline generalises beyond the NSW HSC context to other curricula, jurisdictions, and subject domains. 
The AI marking Pipeline appears to show promise in the area of Economics. It should be tested on other social science subjects such as Legal Studies and Business Studies to assess its efficacy in other subjects where content can change quickly in line with changing social and technological factors affecting society, the law and commercial practices. 

Further areas of experimentation for the Pipeline could include longer response questions: 8 to 10 marks which use directive verbs such as assess. It could also be assessed on 15-mark responses where ``themes and challenges'' rather than outcomes alone would need to be integrated. Moreover, this subject requires significant levels of link to \textit{authority}. Links to authority include cases, laws, media from appropriate sources, documents, international laws and statistics. 20-mark Economics responses, require similar reference to authority with different sources and 20-mark Business Reports require certain writing structures and formats. Legal Studies responses can extend to 25 marks in the options part of the course. 

\subsubsection{Dialogue-based assessment}

Beyond the automation of formative assessment, LLMs enable new forms of interactive assessment. Dialogue-based assessment allows AI systems to probe student understanding through iterative questioning and reflection, resembling human tutor–student interactions and offering greater flexibility for adaptive learning paradigms. Architecturally, this raises challenges related to managing multi-turn context, supporting multimodal input and output, and ensuring that conversational assessment remains aligned with authorised curriculum intent and assessment standards.

\subsubsection{AI–student synergy assessment}

AI-collaboration assessment evaluates how students prompt, verify, and integrate AI-generated responses—capabilities that are increasingly essential in the era of generative technologies. This perspective reframes assessment as a dynamic, context-aware process co-shaped by human learners and AI. From an architectural perspective, such assessment scenarios require explicit modelling of AI usage boundaries, provenance of AI-assisted reasoning, and mechanisms for distinguishing student understanding from AI contribution.

\subsubsection{Formalising tacit knowledge as first-class architectural artefacts}

While this paper embeds tacit knowledge through authorised documents, historical practices, and marking principles, a key research challenge lies in making such tacit judgement more explicit, inspectable, and evolvable over time. Future research will explore representations of tacit judgement as structured constraints, heuristics, or pattern libraries, and examine how these artefacts can be versioned, audited, and adapted as curricula, cohorts, and disciplinary norms evolve. Tacit knowledge is also expected to play a critical role in dialogue-based assessment, where qualitative judgement and interpretive flexibility are essential.



\section{Conclusion}
\label{sec:conclusion}

This paper presented a curriculum-grounded, configurable LLM pipeline for question-level marking in 
educational assessment, moving beyond prompt-based design towards a verifiable and auditable software architecture. By treating authorised curriculum artefacts and marking principles as first-class components, the pipeline embeds professional judgement at design time while supporting scalable, consistent marking at runtime. Through staged LLM workflows, explicit verification points, and controlled configurability, the proposed architecture demonstrates how context engineering  can operationalise responsible generative AI for assessment. Preliminary evaluation shows that the proposed architecture delivers marking outcomes comparable to human tutors, while yielding justifications that are more traceable to authorised curriculum artefacts and marking standards. These evaluation results prove the curriculum grounding and verification are critical architectural mechanisms for trustworthy LLM-as-judge systems.

\bibliographystyle{elsarticle-num}
\bibliography{references}  

\appendix

\subsection{LLM Marking Prompt Template}\label{sec:direct_prompt}
\begin{figure}[th]
\begin{tcolorbox}[
  title={Direct LLM Marking Prompt},
  colback=gray!5,
  colframe=black!60,
  boxrule=0.5pt,
  left=6pt,
  right=6pt,
  top=6pt,
  bottom=6pt,
  width=\columnwidth
]

Using the following context from the Economics syllabus and performance bands:\\
Performance Bands:\{context\_dpbs\} \\
Syllabus:\{context\_syllabus\} \\
Verbs:\{verbs\} \\
Question: \{question\_text\} \\
Answer: \{answer\_text\} \\
Maximum Marks: \{question\_total\_marks\} \\
Marking examples:\{samples\} \\
\\
First predict which HSC learning outcomes in the syllabus the question is tested against based on the verbs used in the question, and then mark the answer according to these outcomes. use the marking examples and performance band descriptions as a reference if available when marking. Return a json string containing the mark as a number, the predicted learning outcomes, and a short justifications of the mark.

\end{tcolorbox}
\end{figure}

\subsection{Alignment Between Marking Guideline Principles and the Proposed Marking Pipeline}
\label{app:marking_guideline}
\begin{table*}[t]
\renewcommand{\arraystretch}{1.3}
\caption{Alignment Between Marking Guideline Principles and the Proposed Marking Pipeline}
\label{tab:nsw_principles_alignment}
\centering
\begin{tabular}{|p{0.2\textwidth}|p{0.7\textwidth}|}
\hline
\textbf{Principle (Summary)} & \textbf{How the Pipeline Meets the Principle (and Gaps)} \\
\hline
Context of relevant syllabus outcomes and content &
Retrieval is grounded in a structured syllabus database with matched outcomes, skills, concepts, and topics. Marking criteria generation is explicitly anchored to outcomes and topics, ensuring curriculum context is preserved throughout criteria construction. \\
\hline
Award marks for achievement of aspects of outcomes addressed &
Criteria are explicitly outcome-linked, with each criterion referencing a specific HSC outcome. The marking process prioritises observable behaviours defined in the criteria when awarding marks. \\
\hline
Reflect the nature and intention of the question, expressed as demanded knowledge and skills &
Directive verb extraction provides an explicit interpretation of task demand. Criteria are defined as observable behaviours aligned with the directive verbs and calibrated to the mark value. \\
\hline
Indicate initial criteria used to award marks &
Generated criteria are treated as the initial marking criteria and are produced as an explicit checklist prior to performance calibration. \\
\hline
Allow less predictable or less defined responses where appropriate &
Criteria are expressed as behaviours rather than fixed model answers, and calibrated quality levels describe performance broadly. The marker prompt allows credit for alternative valid approaches that satisfy the behaviours. \textit{Gap}: an explicit instruction to reward originality or flair is not yet embedded in guideline generation. \\
\hline
Extended responses use language consistent with outcomes and band descriptions &
Performance calibration is constrained using authorised performance band descriptors, ensuring that holistic language remains consistent with official outcome and band terminology. \\
\hline
Incorporate the generic rubric (if provided) together with question specifics &
Question-specific signals (directive verb and matched syllabus topics/sub-topics) are incorporated during criteria generation. \\ 
\hline
Clear, unambiguous, and accessible language &
Structured, schema-conformant outputs are enforced to reduce ambiguity, with a small number of consistently phrased criteria and mark ranges. \textit{Gap}: accessibility depends on LLM wording; an explicit readability constraint or check would further strengthen this principle. \\
\hline
Differentiate higher-order outcomes, with more marks allocated to higher-order performance &
Mark ranges and calibrated quality levels are designed to separate comprehensive, sound, partial, and limited performance. Prompts explicitly require stronger differentiation for higher-order tasks, particularly in extended responses. \\
\hline
Indicate the quality required for each mark or sub-range &
Performance calibration outputs explicit mark ranges paired with quality descriptors (e.g.\ 17--20, 13--16). Range structure is dynamically adjusted based on the maximum mark allocation to preserve meaningful discrimination. \\
\hline
High achievement not defined solely by quantity &
Analytic criteria focus on demonstrated behaviours such as accuracy, linkage, reasoning, use of disciplinary terminology, and integration. The marker prompt emphasises judgement against criteria and bands rather than response length. \\
\hline
Optional questions marked using comparable criteria &
Common generation templates and band-descriptor constraints are applied across questions, yielding structurally comparable marking criteria. \textit{Gap}: full comparability across optional questions still requires human moderation and standard-setting beyond template-level consistency. \\
\hline
Generic-type questions use common guidelines exemplified with contexts or content &
Syllabus matching yields stable strands even when multiple contexts are permitted. Guidelines may include optional illustrative examples. \textit{Gap}: exemplars are currently LLM-generated; a deterministic exemplar strategy (e.g.\ drawing from matched concepts or topics) would improve consistency. \\
\hline
\end{tabular}
\end{table*}

\subsection{Judge Prompt used in Evaluation}
\label{app:judge}

\begin{figure}[h]
\begin{tcolorbox}[
  title={Judge Prompt Used in Evaluation},
  colback=gray!5,
  colframe=black!60,
  boxrule=0.5pt,
  left=6pt,
  right=6pt,
  top=6pt,
  bottom=6pt,
  width=\columnwidth
]

You are an independent assessment judge evaluating two alternative markings of the same student response.\\

Your job is to compare Response A vs Response B objectively, using evidence from the student response and the requirements of the question.\\

Evaluate TWO dimensions separately:\\
1) Mark defensibility\\
2) Justification quality\\

Important rules:\\
- Do NOT assume either response is correct.\\
- Do NOT prefer human or AI outputs by default.\\
- For Mark defensibility, ignore how well the mark is explained and ignore writing quality/tone.\\
- For Justification quality, do not reward verbosity; prefer specific, evidence-based explanations.\\
- Penalize any claims not supported by the student response.\\

Mark defensibility definition:\\
Decide which numerical mark is more reasonably defensible given:\\
- the question text (including its task requirements),\\
- the maximum mark available, and\\
- the evidence present in the student response,\\

Using principles consistent with NESA marking practice. In particular:\\
- marks must be awarded only for evidence demonstrated in the student response,\\
- marks must align with the intention and demand of the question,\\
- higher marks require qualitatively stronger evidence, not simply more content,\\
- marks should discriminate appropriately between different levels of achievement.\\

Justification quality definition:\\
Decide which justification better explains and supports its awarded mark by citing specific evidence from the student response and clearly explaining what is present or missing.\\

Output format (STRICT):\\
Mark defensibility: A | B | Tie\\
Justification quality: A | B | Tie\\

Reasons:\\
- ...\\
- ...\\
Return exactly this format and nothing else.\\

\end{tcolorbox}
\end{figure}

\end{document}